\def\year{2020}\relax
\documentclass[letterpaper]{article} 
\usepackage{aaai20}  
\usepackage{times}  
\usepackage{helvet} 
\usepackage{courier}  
\usepackage[hyphens]{url}  
\usepackage{graphicx} 
\usepackage{graphicx}  
\usepackage{pifont}
\usepackage{booktabs}       
\usepackage{amsfonts}       
\usepackage{nicefrac}       
\usepackage{microtype}      
\usepackage{enumitem}
\usepackage{bbm}
\usepackage{mathtools}

\title{SOGNet: Scene Overlap Graph Network for Panoptic Segmentation}
\author{Yibo Yang\textsuperscript{\rm 1,2,}\thanks{Equal Contribution},
Hongyang Li\textsuperscript{\rm 2,$*$},
Xia Li\textsuperscript{\rm 2,3},
Qijie Zhao\textsuperscript{\rm 4}, 
Jianlong Wu\textsuperscript{\rm 2,5}, 
Zhouchen Lin\textsuperscript{\rm 2,}\thanks{Corresponding Author}\\ 
\textsuperscript{\rm 1}Center for Data Science, Academy for Advanced Interdisciplinary Studies, Peking University\\
\textsuperscript{\rm 2}Key Laboratory of Machine Perception (MOE), School of EECS, Peking University\\
\textsuperscript{\rm 3}Key Laboratory of Machine Perception, Shenzhen Graduate School, Peking University\\
\textsuperscript{\rm 4}Wangxuan Institute of Computer Technology, Peking University\\
\textsuperscript{\rm 5}School of Computer Science and Technology, Shandong University\\
{\tt \{ibo,lhy\_ustb,ethanlee,zhaoqijie,jlwu1992,zlin\}@pku.edu.cn} 
}

\begin{document}

\maketitle

\begin{abstract}
The panoptic segmentation task requires a unified result from semantic and instance segmentation outputs that may contain overlaps. However, current studies widely ignore modeling overlaps. In this study, we aim to model overlap relations among instances and resolve them for panoptic segmentation. Inspired by scene graph representation, we formulate the overlapping problem as a simplified case, named scene overlap graph. We leverage each object’s category, geometry and appearance features to perform relational embedding, and output a relation matrix that encodes overlap relations. In order to overcome the lack of supervision, we introduce a differentiable module to resolve the overlap between any pair of instances. The mask logits after removing overlaps are fed into per-pixel instance \verb|id| classification, which leverages the panoptic supervision to assist in the modeling of overlap relations. Besides, we generate an approximate ground truth of overlap relations as the weak supervision, to quantify the accuracy of overlap relations predicted by our method. Experiments on COCO and Cityscapes demonstrate that our method is able to accurately predict overlap relations, and outperform the state-of-the-art performance for panoptic segmentation. Our method also won the Innovation Award in COCO 2019 challenge.
\end{abstract}

\section{Introduction}
\begin{figure}[t!]
		\includegraphics[width=1\linewidth]{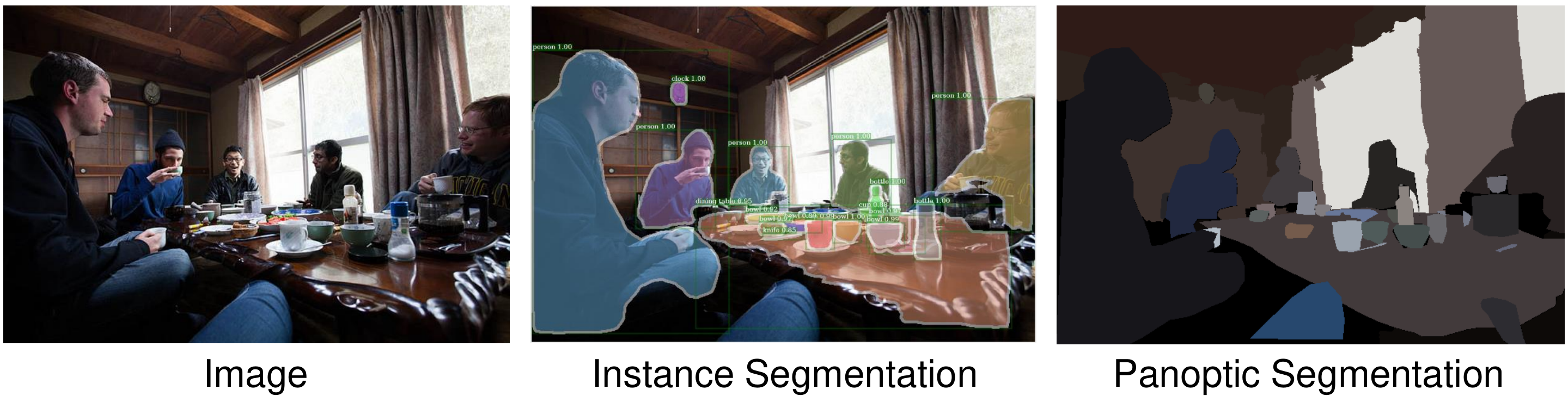}
	\caption{Instance segmentation has overlapping regions for objects, while panoptic segmentation requires a unified result for each pixel. Our study aims to explicitly predict overlap relations and resolve overlaps for the panoptic output.}
	\label{fig1}
\end{figure}

Convolutional Neural Networks (CNNs) have achieved huge success in computer vision tasks such as image recognition \cite{he2016deep,yang2018convolutional}, semantic segmentation \cite{long2015fully,chen2018deeplab}, object detection \cite{girshick2015fast,ren2015faster}, and instance segmentation \cite{he2017mask}. The semantic segmentation task answers which background scene a pixel belongs to, while the instance segmentation task predicts foreground object masks. Recently, the panoptic segmentation task introduced in \cite{kirillov2018panoptic} aims to unify the results of semantic and instance segmentation into a single pipeline. The system performs semantic segmentation for pixels that belong to amorphous background scenes, named \emph{stuff}. For countable foreground objects, named \emph{things}, the goal is to assign each object region with the right thing class, as well as an instance \verb|id|, identifying which object it belongs to. As a result, panoptic segmentation cannot have overlapping segments. However, most cutting-edge high-performance instance segmentation methods \cite{he2017mask} adopt the region-based strategy \cite{girshick2014rich}, and output overlapping segments. As shown in Figure \ref{fig1}, the object pairs, such as \emph{cup}-\emph{dinning table}, \emph{bottle}-\emph{dinning table}, and \emph{bowl}-\emph{dinning table}, share overlapping regions from instance segmentation output. Therefore, resolving overlaps and producing coherent segmentation results are the main challenge for the panoptic segmentation task \cite{kirillov2018panoptic}. 


In \cite{kirillov2018panoptic}, the semantic and instance segmentation are trained separately, and their panoptic results are merged by heuristic post-processing steps. Later studies aim to unify the semantic and instance segmentation into an end-to-end training framework \cite{kirillov2019panoptic,li2018attention,liu2019end,xiong2019upsnet,mapillary,yang2019deeperlab,li2018learning}. The panoptic results are usually produced by fusion strategies \cite{kirillov2019panoptic,li2018attention}, or predicted by a panoptic head \cite{liu2019end,xiong2019upsnet}. These studies do not explicitly model overlap relations among objects, which is especially important for datasets with rich categories and complex scenes. However, modeling overlap is challenging without the supervision of object relations or depth information. 

In this study, we introduce the scene overlap graph network (SOGNet) for panoptic segmentation. The SOGNet consists of four components: the joint segmentation, the relational embedding module, the overlap resolving module, and the panoptic head. The SOGNet trains semantic and instance segmentation in an end-to-end fashion, explicitly encodes overlap relations, resolves the overlap between any pair of objects in a differentiable way, and outputs a unified panoptic result in the panoptic head.

Similar to \cite{kirillov2019panoptic,li2018attention,liu2019end,xiong2019upsnet,mapillary,li2018learning}, we also use ResNets \cite{he2016deep} with feature pyramid network (FPN) \cite{lin2017feature} as the shared backbone for our semantic and instance segmentation branches. Inspired by the relation classification in scene graph parsing tasks \cite{zellers2018neural,woo2018linknet}, we formulate the overlapping problem in panoptic segmentation as a simplified scene graph with directed edges, in which there are only three relation types for instance $i$ with respect to $j$: no overlap, covering as a subject, and being covered as an object. We name this representation as \emph{scene overlap graph} in this study. We leverage the category, geometry, and appearance information of objects to perform edge feature embedding for the scene overlap graph, and output a matrix that explicitly encodes overlap relations. However, different from scene graph parsing tasks with the commonly used Visual Genome dataset \cite{krishna2017visual} that has relation annotations, the panoptic segmentation task does not offer annotations of object relations or depth information, so the overlap relations cannot be modeled with direct supervision.


In order to overcome this problem, we develop the overlap resolving module, which resolves the overlaps between any pair of instances in a differentiable way. The mask logits after removing overlaps are then used for per-pixel instance \verb|id| classification in the panoptic head with the panoptic annotation. In doing so, the supervision from pixel-level classification helps the instance-level modeling of overlap relations.


We list the contributions in this study as follows:
\begin{itemize}
	\item We formulate the overlapping problem in panoptic segmentation as a structured representation, named scene overlap graph. Using category, geometry and appearance features, we perform relational embedding and output a matrix that explicitly encode overlap relations. 
	\item In order to deal with the lack of supervision on overlap relations, we develop an overlap resolving module that resolves overlaps between any pair of instances in a differentiable way. The supervision from per-pixel instance \verb|id| classification in the panoptic head helps to encode overlap relations. We also generate an approximate ground truth as weak supervision to quantify the accuracy of overlap relations predicted by our network. 
	\item Experiments on the COCO and Cityscapes datasets show that, our proposed method is able to accurately predict overlap relations, and outperform the state-of-the-art performance for panoptic segmentation. 
\end{itemize}

\section{Related Work}

\subsubsection{Image Segmentation}


The semantic segmentation task focuses on background scenes and is based on fully convolutional networks (FCNs) \cite{long2015fully}. Because detail information is important for dense prediction problems, later studies learn finer representation by deconvolution \cite{noh2015learning}, encoder-decoder structures \cite{badrinarayanan2017segnet}, or introducing skip connections between down-sampling and up-sampling paths \cite{ronneberger2015u}. Other methods aim to aggregate multi-scale context \cite{farabet2013learning,chen2018deeplab,zhao2017pyramid}, and better capture long-range dependencies \cite{zheng2015conditional,li2019expectation}. The instance segmentation task deals with foreground objects. Similar to object detection \cite{girshick2015fast,ren2015faster}, many instance segmentation studies \cite{li2017fully,he2017mask} also adopt the region-based strategy \cite{girshick2014rich}, and are able to achieve strong performance due to accurate localization for instances. As another stream, segmentation-based methods \cite{liang2018proposal,arnab2017pixelwise} perform pixel-wise classification and then construct object instances by grouping. 

The recently proposed task, panoptic segmentation \cite{kirillov2018panoptic}, requires a unified result for background scenes and foreground objects. A naive implementation is to train the two sub-tasks separately, and then fuse the results by heuristic rules \cite{kirillov2018panoptic}. Follow-up studies train semantic and instance segmentation in an end-to-end network by sharing backbone \cite{de2018panoptic,kirillov2019panoptic,li2018attention,liu2019end,xiong2019upsnet,mapillary,yang2019deeperlab,li2018learning}. Most of them use fusion heuristics to produce the final output. In \cite{liu2019end,xiong2019upsnet}, a panoptic head is constructed to predict instance \verb|id|. Li \emph{et al.} \cite{li2018learning} introduce a binary mask to differentiate between thing or stuff for each pixel. A semi- and weakly-supervised method is proposed in \cite{li2018weakly} to relieve the cost of pixel-level annotation. 

An important aspect ignored by current panoptic segmentation studies is modeling and resolving overlaps. The study \cite{lazarow2019learning} tries to learn instance occlusions but cannot resolve them in the end-to-end training. As a comparison, our study is able to explicitly model overlap relations, telling us whether an instance lies upon or beneath another, and resolve their overlaps in a differentiable way to generate the panoptic output. 

\subsubsection{Relational Modeling}

Parsing relationships of objects has been one of the core components of visual understanding. In \cite{hu2018relation}, appearance and geometry features are used to build interactions for object detection. The visual relationship datasets, such as Visual Genome, inspire a series of studies on scene graph generation. In \cite{zellers2018neural,woo2018linknet}, the low-rank outer product \cite{kim2016hadamard} is adopted to perform relational embedding from object features. Other relation reasoning methods are proposed by graph-based propagation \cite{xu2017scene}, associative embedding \cite{newell2017pixels}, and introducing an efficient module \cite{santoro2017simple}.

\begin{figure*}[!th]
	\vspace{-2mm}
	\centering
		\includegraphics[width=0.84\linewidth]{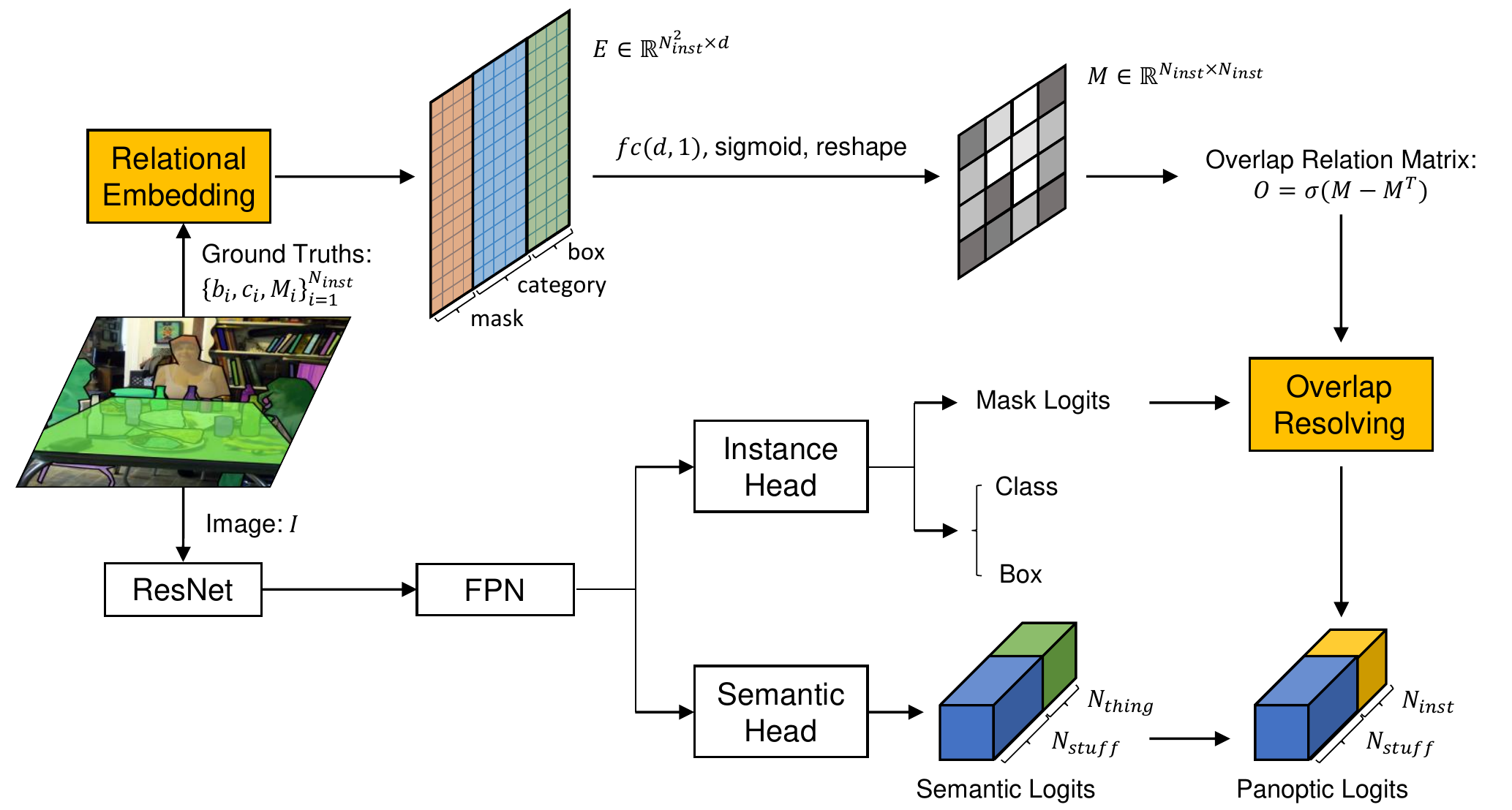}
	\vspace{-2mm}
	\caption{An illustration of the SOGNet for panoptic segmentation. The instance ground truths are input of our relational embedding module. During inference, they are replaced with the predictions from the instance segmentation head. The architecture is trained in an end-to-end manner. $\sigma$ denotes the ReLU non-linear function.}
	\vspace{-2mm}
	\label{fig2}
\end{figure*}

In our study, we formulate the overlapping problem as a simplified scene graph, and also perform relational embedding to encode overlap relations. Our method differs from these studies in that our problem does not offer relation annotation to supervise. We use the supervision from panoptic head to help the modeling of overlap relations.

\section{Scene Overlap Graph Network}

In the scene graph generation task \cite{zellers2018neural,woo2018linknet,xu2017scene}, objects in an image are constructed as a graph and their relations are directed edges. We formulate the overlapping problem in panoptic segmentation as a similar structure, named scene overlap graph (SOG). There are three relation types for instance $i$ with respect to $j$: no overlap, covering as a subject, and being covered as an object. Our proposed SOGNet consists of four components. The joint segmentation connects semantic and instance segmentation in a unified network. The relational embedding module explicitly encodes overlap relations of objects. After the overlap resolving module, overlaps among instances are removed in a differentiable way. Finally, the panoptic head performs per-pixel instance \verb|id| classification. An illustration of our SOGNet architecture is shown in Figure \ref{fig2}.

\subsection{Joint Segmentation}

Following current popular methods, we use ResNet with FPN as the shared backbone of semantic and instance segmentation branches. The Mask R-CNN structure is adopted for our instance segmentation head, which outputs the box regression, class prediction, and mask segmentation for foreground objects. As for semantic head, the FPN feature maps first go through three $3\times3$ deformable convolution layers \cite{dai2017deformable}, and then are up-sampled to the $1/4$ scale. Finally, they are concatenated to generate the per-pixel category prediction. This branch is supervised with both stuff and thing classes, and then the semantic logits of stuff classes are extracted into the panoptic head. We train our model using instance and panoptic annotation. The panoptic annotation that gives per-pixel category and instance \verb|id| supervises the semantic and panoptic head, respectively. The instance annotation contains overlaps and is used for instance segmentation.

\subsection{Relational Embedding Module}

For any training image, we are given the ground truth $\{b_i,c_i,M_i\}_{i=1}^{N_{inst}}$, where $b_i$, $c_i$, and $M_i$ refer to the bounding box, one-hot category, and corresponding mask for instance $i$, respectively, and $N_{inst}$ is the number of instances in this image. As illustrated in Figure \ref{fig2}, we perform relational embedding using the ground truth in the training phase. During inference, we replace them with the prediction from Mask R-CNN branch. The $b_i\in\mathcal{R}^4$ and $c_i\in\mathcal{R}^{80}$ (there are 80 thing classes for COCO) encode geometry and category information, respectively. In order to include appearance feature, we resize the values inside box $b_i$ from $M_i$ as $28\times28$, which is consistent with the size of Mask R-CNN's output. The resized mask is flattened to be a vector, denoted as $m_i\in\mathcal{R}^{784}$.

The bilinear pooling method learns joint representation for pair of features and is widely applied to visual question answering \cite{kim2016hadamard,kim2018bilinear}, and image recognition \cite{yu2018hierarchical} tasks. We construct our category and appearance relation features using the low-rank outer product in \cite{kim2016hadamard}. For a pair of instances $i$ and $j$, their category relation feature is calculated as: 
\begin{equation}
E^{(c)}_{i|j}=P^T\left(\sigma(V^Tc_i)\circ \sigma(U^Tc_j)\right),
\label{eq1}
\end{equation}
where $\circ$ denotes the Hadamard product (element-wise multiplication), $\sigma$ is the ReLU non-linear activation, $V$ and $U$ are two linear embeddings that project the input into subject and object features, respectively, and $P$ maps the relation feature into output dimension $d_c$. We then have the category relation features as:
\begin{equation}
E^{(c)}=\left[E^{(c)}_{1|1},E^{(c)}_{1|2},\cdots,E^{(c)}_{N_{inst}|N_{inst}}\right]^T\in\mathcal{R}^{N^2_{inst}\times d_c},
\end{equation}
where ``[\ ]'' is the concatenation operation. In a similar way, using $m_i$ as the input of Eq. (\ref{eq1}), we can also construct the appearance relation features $E^{(m)}\in\mathcal{R}^{N^2_{inst}\times d_m}$.

The relative geometry provides strong information to infer whether two objects have overlap or not. Following \cite{hu2018relation,woo2018linknet}, we have the translation- and scale-invariant relative geometry feature encoded as:
\begin{equation}
E^{(b)}_{i|j}=K^T\left(\frac{x_i-x_j}{w_j}, \frac{y_i-y_j}{h_j}, \log\left(\frac{w_i}{w_j}\right), \log\left(\frac{h_i}{h_j}\right)\right)^T,
\end{equation}
where $x_i,y_i,w_i,h_i$ are coordinates and scales extracted from $b_i$, and $K\in\mathcal{R}^{4\times d_b}$ is a linear matrix that maps the 4-dimensional relative geometry feature into high-dimensional $d_b$. We can further have the geometry relation features $E^{(b)}\in\mathcal{R}^{N^2_{inst}\times d_b}$. We concatenate these edge representations about appearance, category, and geometry as: 
\begin{equation}
E=[E^{(m)}, E^{(c)}, E^{(b)}]\in\mathcal{R}^{N^2_{inst}\times d},
\end{equation}
where $d=d_m+d_c+d_b$. The relational embedding is further used to encode overlap relations.

\subsection{Overlap Resolving Module}

Based on relational embedding, we introduce the overlap resolving module to explicitly model overlap relations and resolve overlaps among instances in a differentiable way. 

As illustrated in Figure \ref{fig2}, the relation features, $E\in\mathcal{R}^{N^2_{inst}\times d}$, go through a $fc(d,1)$ layer to have a single-channel output with the sigmoid activation to restrict the values within $(0,1)$. We reshape the output as a square matrix, denoted as $M\in\mathcal{R}^{N_{inst}\times N_{inst}}$. The element $M_{ij}$ has a physical meaning to represent the potential of instance $i$ being covered by instance $j$. Because there can be only one overlap relation between instances $i$ and $j$, we then introduce the overlap relation matrix defined as:
\begin{equation}
O=\sigma(M-M^T)\in\mathcal{R}^{N_{inst}\times N_{inst}},
\end{equation}
where $\sigma$ denotes the ReLU activation that is used to filter out the negative differences between potentials on symmetric positions. In doing so, if $O_{ij}>0$, it encodes that instance $i$ is being covered by instance $j$, while on its symmetric position, $O_{ji}=0$. When $O_{ij}=O_{ji}=0$, the instances $i$ and $j$ do not have overlaps. Besides, all diagonal elements $O_{ii}$ equals to 0. As explained later, the positive elements in $O$ will be optimized towards $1$ in implementations. We now show how to leverage the overlap relation matrix $O$ to resolve overlaps. 

For each bounding box, $b_i$, of the ground truth instances, we have its mask logits (the activations before sigmoid) of $28\times 28$ from the Mask R-CNN output. We then interpolate these logits back to the image scale $H\times W$ by bilinear interpolation and padding outside the box. These logits, denoted as $\{A_i\}_{i=1}^{N_{inst}}$, may have overlaps because Mask R-CNN is region-based and operates on each region independently. Using the matrix $O$, we can deal with the overlaps between instances $i$ and $j$ as:
\begin{equation}
A'_i=A_i-A_i\circ\left[s(A_i)\circ s(A_j)\right]O_{ij},
\label{eq6}
\end{equation}
where $A'_i$ is the output logit of instance $i$, and $s$ represents the sigmoid activation that turns the logit $A_i$ into a binary-like mask $s(A_i)$. The element-wise multiplication, $s(A_i) \circ s(A_j)$, calculates the intersecting region between instances $i$ and $j$. The value $O_{ij}$ decides whether the elements in intersecting region should be removed from the logit $A_i$. When $O_{ij}$ approaches $1$, $O_{ji}$ equals to $0$, thus the logit $A_j$ will not be affected, and vice versa. 

Considering the overlap relations of all the other instances on $i$, we have:
\begin{equation}
A'_i=A_i-A_i\circ s(A_i) \circ \sum_{j=1}^{N_{inst}}s(A_j) O_{ij},
\end{equation}
and then the computational steps of the overlap resolving module can be formulated as:
\begin{equation}
\mathcal{A}'=\mathcal{A}-\mathcal{A}\circ s(\mathcal{A}) \circ \left(s(\mathcal{A})\times_3 O^T\right),
\label{eq8}
\end{equation}
where $\mathcal{A}=[A_1,A_2,\cdots,A_{N_{inst}}]\in\mathcal{R}^{H\times W\times N_{inst}}$, and $\times_3$ denotes the Tucker product along the 3-rd dimension (reshape $s(\mathcal{A})$ as $\mathcal{R}^{HW\times N_{inst}}$ for inner product with $O^T$, and then return to $\mathcal{R}^{H\times W\times N_{inst}}$). We see that our module is friendly to tensor operations in current deep learning frameworks, and is differentiable for resolving overlaps, so that the SOGNet can be trained in an end-to-end fashion.

\subsection{Panoptic Head}

The overlap relation matrix, $O$, explicitly encodes whether there is intersection between any pair of instances, and if there is, the overlapping region should be removed from which instance. However, we are not provided with the supervision of overlap relations by the panoptic segmentation task. Because accurately resolving overlaps has a strong correlation with the quality of final panoptic output, we can exploit the pixel-level panoptic annotation to assist in the process of modeling overlap relations encoded by $O$. As illustrated in Figure \ref{fig2}, the instance logits $\mathcal{A'}$ after the SOG module are then fed into the panoptic head. 

Following UPSNet \cite{xiong2019upsnet}, we incorporate the logits from semantic head into the mask logits $\mathcal{A'}$. We get the logits of $i$-th object from semantic output $X_i$ by taking the values inside its ground truth box $B_i$ from the channel corresponding to its ground truth category $C_i$, and padding zeros outside the box. In UPSNet, they are combined by addition, which is denoted as ``Panoptic Head 1''. Here we develop an improved combination denoted as ``Panoptic Head 2''. They are compared as:
\begin{align}
& {\rm Panoptic\ Head\ 1}:\quad Z_i=X_i+A'_i, \\
& {\rm Panoptic\ Head\ 2}:\quad Z_i=k \cdot X_i \circ s(A'_i) + A'_i,
\end{align}
where $Z_i$ is the combined logit, $s$ denotes the sigmoid function and $k$ is a factor to balance the numerical difference between semantic output values and mask logits. We set $k$ to be 2 in our experiments. Finally, we concatenate the combined instance logits $\mathcal{Z}_{inst}=[Z_1,...,Z_{N_{inst}}]$ and the stuff logits $\mathcal{Z}_{stuff}$ from the semantic head to perform per-pixel instance \verb|id| classification with the standard cross entropy loss function, $\mathcal{L}_{panoptic}$.


Despite we do not have the supervision to know which instance lies on the other one, we can leverage the ground truth binary masks, $\{M_i\}_{i=1}^{N_{inst}}$, to infer whether two instances have overlaps or not. We produce a symmetric relation matrix $R\in\mathcal{R}^{N_{inst}\times N_{inst}}$ defined as:
\begin{equation}
R_{ij}= \mathbbm{1}\left[\frac{|M_i\circ M_j|}{\min\{|M_i|,|M_j|\}}\ge 0.1\right],\quad i\ne j,
\end{equation}
where $|\cdot|$ calculates the area of a binary mask through sum operation, $\circ$ calculates the intersection mask through element-wise multiplication, and $\mathbbm{1}$ denotes the indicator function that equals to 1 when the condition holds. All diagonal elements $R_{ii}$ are filled with $0$. When $R_{ij}=R_{ji}=1$, it indicates that the overlapped intersection over the smaller object is larger than $0.1$, which means there is a significant overlap between instances $i$ and $j$. With the symmetric relation matrix $R$, we can introduce the relation loss function as:
\begin{equation}
\mathcal{L}_R=\frac{1}{N^2_{inst}}\left\|O+O^T-R\right\|^2_F,
\label{eq10}
\end{equation}
which calculates the mean squared error between $(O+O^T)$ and $R$. In doing so, when there is overlap between instances $i$ and $j$, \emph{i.e.}, $R_{ij}=R_{ji}=1$, the overlap relation $O_{ij}$ or $O_{ji}$ is forced to approach $1$, so that it will not contribute trivially when removing overlaps by Eq. (\ref{eq6}). 

In total, our SOGNet has the loss functions for semantic and instance segmentation, the panoptic loss $\mathcal{L}_{panoptic}$ for instance \verb|id| classification, and the relation loss $\mathcal{L}_{R}$ to help optimizing the overlap relation matrix $O$. 

\subsection{Evaluation Metrics}

\subsubsection{Panoptic Quality}

We adopt the evaluation metric introduced in \cite{kirillov2018panoptic}, called Panoptic Quality (PQ). It can be viewed as the multiplication of a segmentation term (SQ) and a recognition term (RQ):
\begin{equation}
{\rm PQ}=\underbrace{\frac{\begin{matrix}
		\sum_{(p,q)\in TP} {\rm IoU}(p,g) 
		\end{matrix}}{|TP|}}_{\rm SQ}
\times
\underbrace{\frac{|TP|}{|TP|+\frac{1}{2}|FP|+\frac{1}{2}|FN|}}_{\rm RQ},
\end{equation}
where $p$ and $g$ are predicted and ground truth segments, and $TP$, $FP$ and $FN$ denote the true positive, false positive and false negative sets, respectively. 

\subsubsection{Overlap Accuracy}

For dataset, such as COCO, the instance annotation permits overlapping instances, while the panoptic annotation contains no overlaps. We can leverage the difference between the two annotations to generate an approximate ground truth of overlap relations, in order to test the quality of overlap relations predicted by our model. The method is also used in \cite{lazarow2019learning} to generate their occlusion ground truth. 

Concretely, we are provided with the instance annotation $\{M_i\}^{N_{inst}}_{i=1}$ , and the panoptic annotation $\{\hat{M}_i\}^{N_{inst}}_{i=1}$. For any pair of instances $i$ and $j$, we calculate the intersecting region by $M_i\circ M_j$, and inspect which one of $\hat{M}_i$ and $\hat{M}_j$ mainly covers the intersecting region, to know if $i$ lies upon $j$ or the other way round. Note that the instance and panoptic annotation are not seamlessly matched. Thus this method can only produce approximately true overlap relations

Using the synthetic ground truth as weak supervision, we construct a new asymmetric relation matrix $R^{\star}$. When $R^{\star}_{ij}=1$, we have $R^{\star}_{ji}=0$, and it means instance $i$ is covered by $j$. We can have a new relation loss function in this weakly-supervised setting to replace Eq. (\ref{eq10}) with:
\begin{equation}
\mathcal{L}^{\star}_{R}=\frac{1}{N^2_{inst}}\left\|O-R^{\star}\right\|^2_F,
\label{eq11}
\end{equation}
which directly supervises the overlap relation matrix $O$. In experiments, the weakly-supervised manner by Eq. (\ref{eq11}) and our method by Eq. (\ref{eq10}) have similar performances. Note that the supervision is only valid for datasets such as COCO that has difference between instance and panoptic annotations. It will be ineffective for datasets such as Cityscapes. But our method by Eq.(\ref{eq10}) works in both cases. 

Thus the weakly-supervised manner by Eq. (\ref{eq11}) is served to test the efficacy of our method. Using the weak supervision $R^{\star}$, we develop a metric, named overlap accuracy (OA), to quantify the quality of overlap predictions encoded by $O$. The OA of image $I$ is calculated as:
\begin{equation}
{\rm OA}(I)=\frac{|TP|+|TN|}{N_{inst}\times (N_{inst}-1)},
\end{equation}
where $TP=\{(i,j)|R^{\star}_{ij}=1, O_{ij}\ge 0.5, i\ne j\}$, and $TN=\{(i,j)|R^{\star}_{ij}=0, O_{ij}<0.5, i\ne j\}$. Our reported OA is an average over all images in the validation set. 

\section{Experiments}

We conduct experiments on the COCO and Cityscapes datasets for panoptic segmentation, and show that our proposed SOGNet is able to accurately predict overlap relations and outperform state-of-the-art performances.


\subsection{Implementation Details}

\subsubsection{Training}

\begin{figure*}[ht!]
	\centering
		\includegraphics[width=0.98\linewidth]{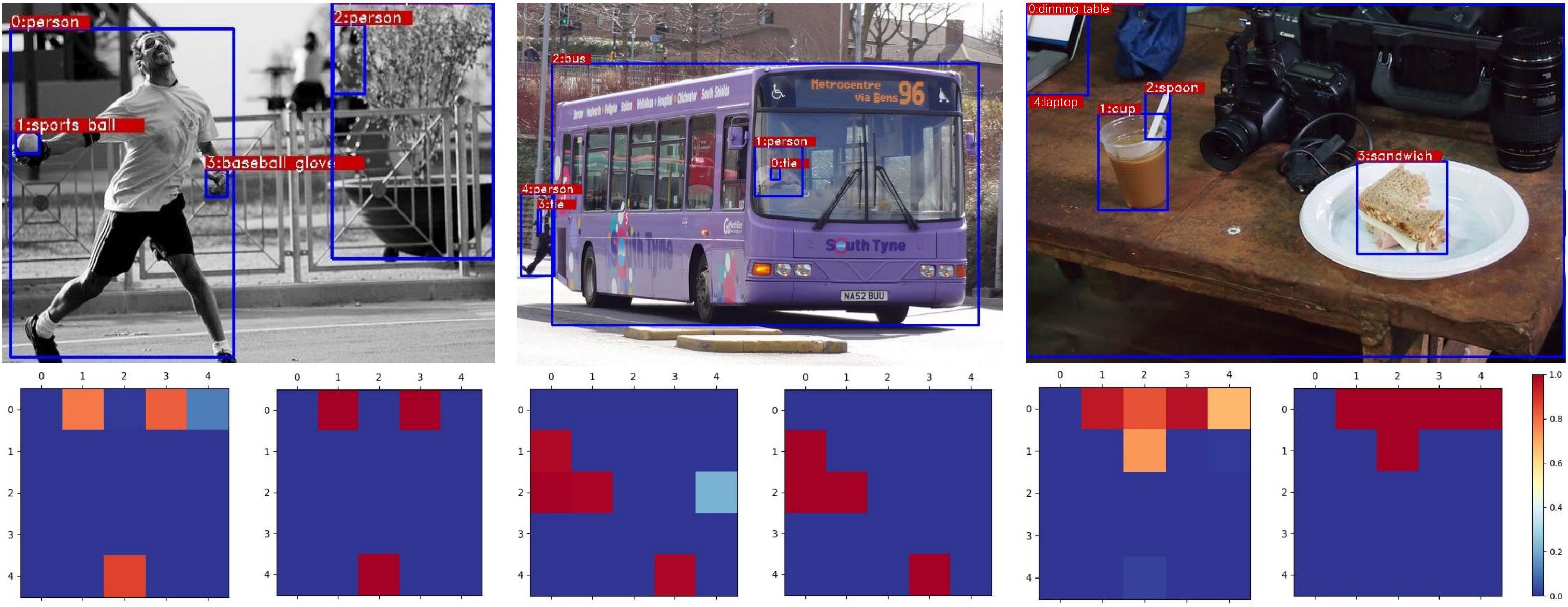}
	\vspace{-0.5mm}
	\caption{Visualization of the overlap relations encoded by $O$ (down left) and the approximate ground truth, $R^{\star}$ (down right). Note that the activation on location $(i,j)$ represents that the instance $i$ is covered by (lies below) $j$. The indices of instances are marked in the images. Zoom in to have a better view. More visualization results can be found in the supplementary material.}
	\label{overlap}
	\vspace{-1mm}
\end{figure*}

We set the weights of loss functions following \cite{xiong2019upsnet}. The weight of panoptic head is 0.1 for COCO and 0.5 for Cityscapes. The weight of relation loss is set to 1.0. We train the models with a batchsize of $8$ images distributed on 8 GPUs. The region proposal network (RPN) is also trained end-to-end. The SGD optimizer with 0.9 Nesterov momentum and a weight decay of $10^{-4}$ is used. We use an equivalent setting to UPSNet for fair comparison. Images are resized with the shorter edge as 800, and the longer edge less than 1333. We freeze all batch normalization (BN) \cite{ioffe2015batch} layers within the ResNet backbone. For COCO, we train the SOGNet for 180K iterations. The initial learning rate is set to $0.01$ and is divided by $10$ at the 120K-th and 160K-th iterations. For Cityscapes, we train for 24K iterations and drop the learning rate at the 18K-th iteration. Besides, in order to test the quality of our overlap predictions, we perform an ablation study on COCO using a shorter training schedule because our relation loss converges soon. We only train for 45K iterations and drop the learning rate at iteration 30K and 40K. We do not adopt the void channel prediction proposed in UPSNet. In implementations, we filter out the instances that have no overlap with any other instance to reduce negative samples and computation overhead. 


\subsubsection{Inference}

During inference phase, the  ground truths $\{b_i,c_i,M_i\}_{i=1}^{N_{inst}}$ as the input of our relational embedding are replaced with the predictions from Mask R-CNN branch. In order to remove invalid instances, we filter out instances whose probability is lower than a threshold, and perform an NMS-like procedure, following \cite{kirillov2018panoptic,xiong2019upsnet}. For highly overlapped predictions of the same class, we keep the mask with the higher confidence score and discard the other one if the intersection is larger than a threshold. Otherwise, we keep the non-interacting part and deal with the next instance. The final output is predicted by our panoptic head. For stuff segment whose area is lower than 4096, we set the corresponding region as void. 

\subsection{Ablation Study}

\begin{table}[!t]
	\renewcommand\arraystretch{1.1}
	\centering
	\begin{tabular}{ccc|ccc|c}
		\hline
		Box & Cat & Mask & PQ & SQ & RQ & OA\\
		\hline\hline
		\ding{51} & - & - & 37.5 & 76.3 & 47.0 & 69.62\\
		- & \ding{51} & - & 37.9 & 76.6 & 47.3 & 75.48\\
		\ding{51} & \ding{51} & - & 38.3 & 76.8 & 47.7 & 88.19\\
		\ding{51} & \ding{51} & \ding{51} & \textbf{38.4} & \textbf{76.9} & \textbf{47.8} & \textbf{89.22}\\
		\hline
		\multicolumn{3}{c|}{Weakly supervised} & 38.4 & 77.0 & 47.7 & 89.31\\
		\hline
	\end{tabular}
	\caption{Different input for the relational embedding module. ``Cat'', ``Box'' and ``Mask'' denote the category, geometry and appearance features, respectively.}
	\label{ablaion}
	\vspace{-1mm}
\end{table}

\begin{table}[!t]
	\renewcommand\arraystretch{1.1}
	\centering
	\begin{tabular}{l|ccc}
		\hline
		Methods & PQ & SQ & RQ \\
		\hline
		PlainNet + heuristics & 39.6 & 78.7 & 48.4 \\
		PlainNet + heuristics + label prior & 40.9 & 78.8 & 49.7 \\
		PlainNet + PH1 & 42.3 & 78.6 & 52.1 \\
		SOGNet (PH1) & 43.0 & 78.1 & 53.1 \\
		SOGNet (PH2) & \textbf{43.7} & 78.7 & \textbf{53.5} \\
		\hline
	\end{tabular}
	\caption{PlainNet denotes the joint segmentation component of SOGNet. They are trained in the same condition. ``PH 1 / 2'' denotes the ``Panoptic Head 1 / 2'', respectively.}
	\label{ablaion2}
	\vspace{-1mm}
\end{table}

We use ResNet-50 as backbone with a short training schedule, and conduct experiments to analyze feature combinations for our relational embedding, and test the quality of overlap relations predicted by our method. As shown in Table \ref{ablaion}, we use different features as the input of our relational embedding. When only category or geometry feature is adopted, the performance improvement on PQ is not so significant, and the overlap prediction does not show high accuracy. When category and geometry features are used together, the embedding becomes much more powerful. Mask feature also slightly improves the overlap accuracy. We expect that a more sophisticated feature design will further boost the performance.  It is observed that the weakly-supervised method by Eq. (\ref{eq11}) achieves a similar result to our method by Eq. (\ref{eq10}). As shown in Figure \ref{overlap}, we visualize the overlap relations predicted by $O$, as well as the approximate ground truth, $R^{\star}$, on images from the validation set. More examples can be found in the supplementary material. It is shown that the matrix $O$ accurately predicts some overlap relations, including \emph{baseball glove}$\rightarrow$\emph{person}, \emph{tie}$\rightarrow$\emph{person}$\rightarrow$\emph{bus}, and \emph{spoon}$\rightarrow$\emph{cup}$\rightarrow$\emph{dinning table}. The results demonstrate that the overlap relations are modeled well with the help of supervision from per-pixel instance \verb|id| classification in the panoptic head. Our method is able to encode overlap relations without direct supervision on them. 

\begin{figure}
	\centering
		\includegraphics[width=1\linewidth]{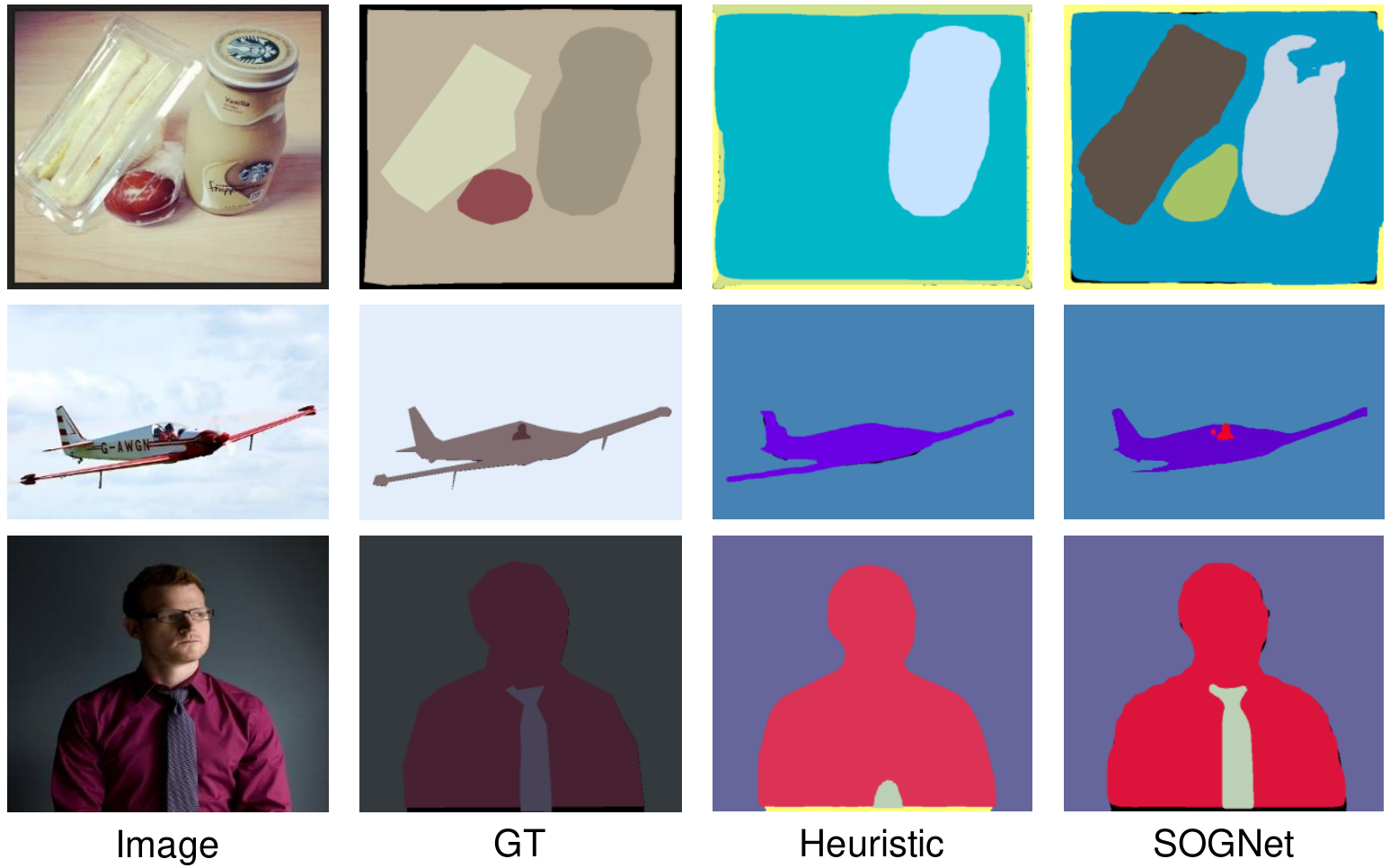}
	\caption{The Visualization of panoptic segmentation results of heuristic inference and SOGNet.}
	\label{panoptic}
\end{figure}

Using the standard training schedule and ResNet-50 as the backbone, we also perform comparisons between SOGNet and heuristic inference. The heuristics in \cite{kirillov2018panoptic} sort instances according to their objectness scores to deal with overlaps. In \cite{li2018attention}, some hand-crafted label priors are made to rule overlap orders. For example, \emph{tie} should always cover \emph{person}. As a comparison, SOGNet explicitly predict overlap relations and resolve overlaps in a differentiable way. We train the joint segmentation component of SOGNet as a PlainNet, and perform inference with different methods. As shown in Table \ref{ablaion2}, label prior helps to improve the performance. When PlainNet adds the panoptic head for inference to produce the panoptic results, the performance becomes better. The SOGNet with relational embedding and overlap resolving has a further improvement. And our proposed Panoptic Head 2 (PH2) performs better than PH1. In Figure \ref{panoptic}, we visualize the panoptic segmentation results of heuristic inference and SOGNet. It is shown that SOGNet better deals with the overlapping problem.


\begin{table}
	\renewcommand\arraystretch{1.25}
	\centering
		\begin{tabular}{l|c|ccc}
			\hline
			Models & backbone & PQ & SQ & RQ \\
			\hline
			Megvii & ensemble & 53.2 & 83.2 & 62.9\\
			Caribbean & ensemble & 46.8 & 80.5 & 57.1\\
			PKU-360 & ResNeXt-152-FPN & 46.3 & 79.6 & 56.1\\
			\hline
			Panoptic FPN & ResNet-101-FPN & 40.9 & - & -\\
			OANet & ResNet-101-FPN & 41.3 & - & -\\
			AUNet & ResNeXt-152-FPN & 46.5 & 81.0 & 56.1\\
			UPSNet & ResNet-101-FPN$^*$ & 46.6 & 80.5 & 56.9\\
			\hline
			SOGNet & ResNet-101-FPN$^*$ & \textbf{47.8} & 80.7 & \textbf{57.6}\\
			\hline
		\end{tabular}
	\caption{Comparisons with SOTA performances on COCO \emph{test-dev} set. The first block shows the top-3 entries in public leaderboard of COCO 2018 competition. The second block shows results in recent literatures. $*$ denotes that the backbone has extra deformable convolution layers and longer training schedule is adopted.}
	\label{cocotest}
\end{table}

\subsection{Comparison with Other Methods}


We run SOGNet on the COCO and Cityscapes datasets, and compare the results with state-of-the-art methods including the method in \cite{li2018weakly}, JSIS \cite{de2018panoptic}, TASCNet \cite{li2018learning}, Panoptic FPN \cite{kirillov2019panoptic}, OANet \cite{liu2019end}, AUNet \cite{li2018attention}, UPSNet \cite{xiong2019upsnet}, and OCFusion \cite{lazarow2019learning}.

As shown in Table \ref{cocotest}, with ResNet-101-FPN as the backbone, our proposed SOGNet achieves the highest single-model performance on the COCO \emph{test-dev} set. It has a 1.3\% PQ improvement than AUNet that uses a larger backbone. SOGNet also performs better than UPSNet using the same backbone and training schedule.  



The results of SOGNet on the COCO and Cityscapes validation set are shown in Table \ref{cityandcoco}. It is observed that SOGNet generalizes well to Cityscapes. It has a 0.7\% improvement than TASCNet and UPSNet. On the COCO validation set, SOGNet has a 1.2\% improvement than UPSNet using the same backbone. The mIoU and AP of SOGNet are 54.56 and 34.2 on COCO, which are similar to the results of UPSNet (54.3 and 34.3 as reported). It indicates that our better panoptic performance is not derived from a stronger semantic or instance segmentation model. More importantly, SOGNet is the only method that can explicitly encode overlap relations and tell us which instance lies upon or beneath another.

\begin{table}[!t]
	\renewcommand\arraystretch{1.2}
	\centering
		\begin{tabular}{l|c|c c c}
			\hline
			Models & backbone & PQ & PQ$^{Th}$ & PQ$^{St}$ \\
			\hline
			\hline
			\multicolumn{5}{c}{Cityscapes}\\
			\hline
			Q.Li \emph{et al.} & ResNet-101 & 53.8 & 42.5 & 62.1 \\
			Panoptic FPN & ResNet-101 & 58.1 & 52.0 & 62.5 \\ 
			TASCNet & ResNet-50 & 59.3 & 56.3 & 61.5 \\
			UPSNet & ResNet-50 & 59.3 & 54.6 & 62.7 \\
			\hline
			SOGNet & ResNet-50 & \textbf{60.0} & \textbf{56.7} & 62.5 \\
			\hline
			\hline
			\multicolumn{5}{c}{COCO}\\
			\hline
			JSIS & ResNet-50 & 26.9 & 29.3 & 23.3 \\
			Panoptic FPN & ResNet-101 & 40.3 & 47.5 & 29.5\\
			OCFusion & ResNet-50 & 41.2 & 49.0 & 29.0 \\
			UPSNet & ResNet-50 & 42.5 & 48.5 & 33.4\\
			\hline
			SOGNet & ResNet-50 & \textbf{43.7} & \textbf{50.6} & 33.2\\
			\hline
		\end{tabular}
	\caption{Panoptic segmentation results of SOGNet and other state-of-the-art methods on Cityscapes and COCO. Multi-scale testing and flipping are not used.} 
	\label{cityandcoco}
\end{table}

\section{Conclusion}

In this study, we aim to model overlap relations and resolve overlaps in a differentiable way for panoptic segmentation. We develop the SOGNet composed of the joint segmentation, the relational embedding module, the overlap resolving module, and the panoptic head. It is able to explicitly encode overlap relations without direct supervision on them. Ablation studies detach SOGNet and analyze the efficacy of each component. Experiments demonstrate that SOGNet accurately predicts overlap relations, and outperforms the state-of-the-art methods on both COCO and Cityscapes.  

\section{Acknowledgements}

Z. Lin is supported by NSF China under grant no.s 61625301 and 61731018 and Zhejiang Lab.

\section{Supplementary Material}

\begin{figure*}[]
	\begin{center}
		\includegraphics[width=1\linewidth]{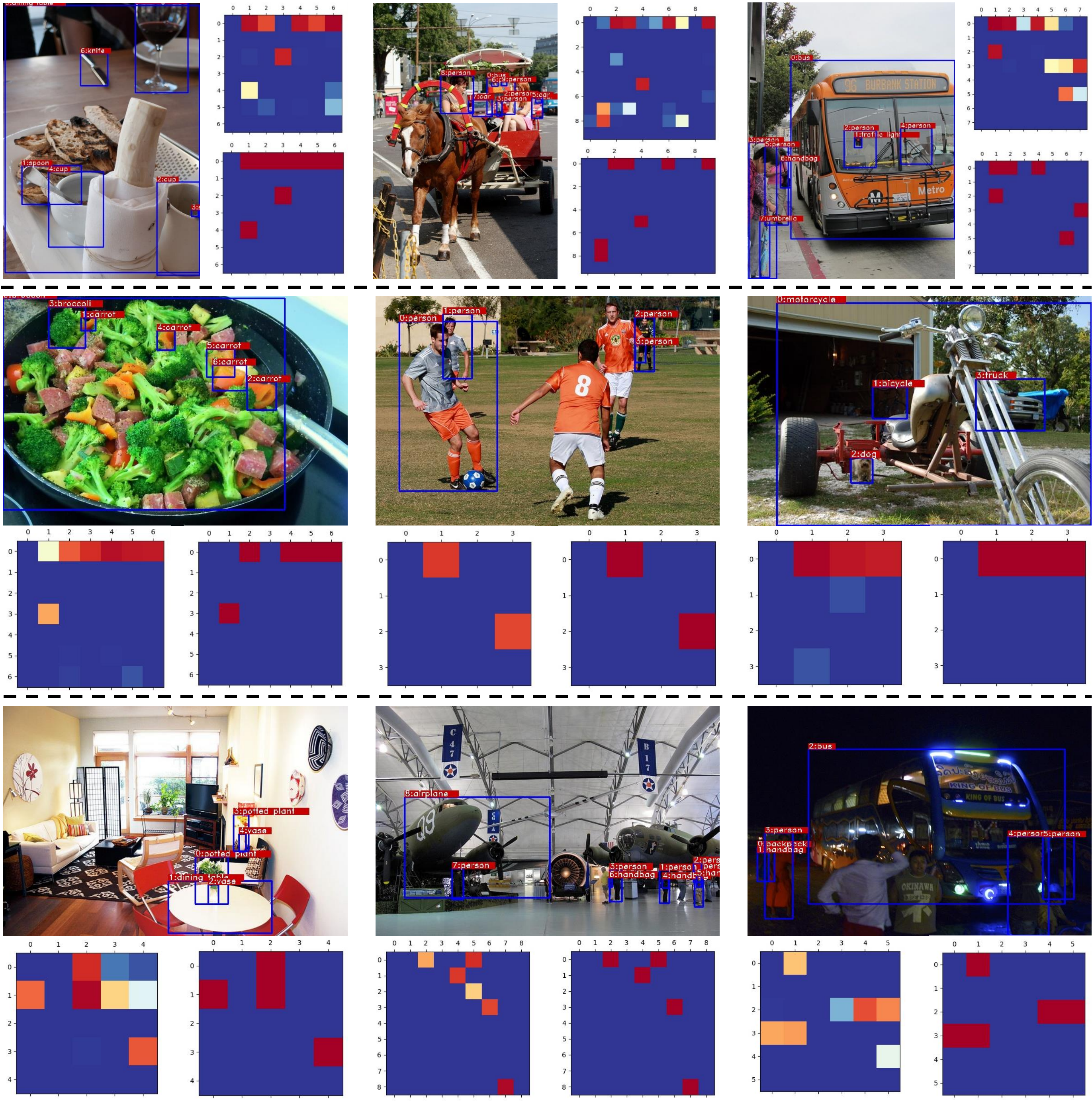}
	\end{center}
	\caption{More visualization examples of the overlap relations predicted from $O$ in SOGNet, and their corresponding approximate ground truths, $R^{\star}$. Note that the activation on location $(i,j)$ represents that instance $i$ is covered by (lies below) instance $j$. The indices of instances are marked in the images. Zoom in to have a better view.}
	\label{panoptic_supp}
	\vspace{-3mm}
\end{figure*}

We offer more visualization examples of the overlap relations predicted by our method, as shown in Figure \ref{panoptic_supp}.

\newpage

\newpage

\bibliographystyle{aaai}\bibliography{sog_aaai.bib}

\end{document}